\documentclass[conference]{IEEEtran}
\pdfoutput=1
\usepackage{cite}
\usepackage{amsmath,amssymb,amsfonts}
\usepackage{algorithm}  
\usepackage{graphicx}
\usepackage{algpseudocode}  
\usepackage[table]{xcolor}
\usepackage{booktabs}
\usepackage[warn]{textcomp}

\usepackage{xcolor}
\def\BibTeX{{\rm B\kern-.05em{\sc i\kern-.025em b}\kern-.08em
    T\kern-.1667em\lower.7ex\hbox{E}\kern-.125emX}}
\begin{document}

\title{Adapted Tree Boosting for Transfer Learning}
\author{\IEEEauthorblockN{1\textsuperscript{st} 
Wenjing Fang}
\IEEEauthorblockA{\textit{Ant Financial Services Group} \\
Hangzhou, China \\
bean.fwj@antfin.com}

\and
\IEEEauthorblockN{2\textsuperscript{nd} 
Chaochao Chen}
\IEEEauthorblockA{\textit{Ant Financial Services Group}\\
Hangzhou, China \\
chaochao.ccc@antfin.com}
\and
\IEEEauthorblockN{3\textsuperscript{nd} 
Bowen Song}
\IEEEauthorblockA{\textit{Ant Financial Services Group}\\
Shanghai, China \\
bowen.sbw@antfin.com}
\and
\IEEEauthorblockN{4\textsuperscript{rd}
 Li Wang}
\IEEEauthorblockA{\textit{Ant Financial Services Group } \\
Hangzhou, China \\
raymond.wangl@antfin.com}
\and
\IEEEauthorblockN{ 5\textsuperscript{th} 
Jun Zhou}
\IEEEauthorblockA{\textit{Ant Financial Services Group} \\
Beijing, China  \\
jun.zhoujun@antfin.com}
\and 
\IEEEauthorblockN{6\textsuperscript{th} 
Kenny Q. Zhu}
\IEEEauthorblockA{\textit{Shanghai Jiao Tong University} \\
Hangzhou, China  \\
kzhu@cs.sjtu.edu.cn}

}
\IEEEoverridecommandlockouts
\IEEEpubid{\makebox[\columnwidth]{978-1-5386-5541-2/18/\$31.00~\copyright2018 IEEE \hfill} \hspace{\columnsep}\makebox[\columnwidth]{ }}
\maketitle
\IEEEpubidadjcol
\begin{abstract}
Secure online transaction is an essential task for e-commerce platforms. Alipay, one of the world's leading cashless payment platform, 
provides the payment service to both merchants and individual customers.
The fraud detection models are built to protect the customers, but stronger demands are raised by the new scenes, which are lacking in training data and labels. 
The proposed model makes a difference by utilizing the data under similar old scenes and the data under a new scene is treated as the target domain to be promoted.
Inspired by this real case in Alipay, we view the problem as a transfer learning problem and design a set of revise strategies to transfer the source domain models to the target domain under the framework of gradient boosting tree models. This work provides an option for the cold-start and data-sharing problems.
\end{abstract}
\begin{IEEEkeywords}
fraud detection, transfer learning, fine tuning, gradient boosting tree
\end{IEEEkeywords}

\section{Introduction}
Two cousins prepared masks and weapons to rob some money. However, they broke into 3 convenience stores before getting caught, only to got 1700 yuan. They are despairing that the money cannot even afford their travelling expenses and tools. This is a real criminal case in Hangzhou,China, a cashless city. The mobile payment spreads all over the city where can pay for almost everything without a wallet. 

The robberies nowadays take place more online and harm the public interest. Alipay\footnote{https://global.alipay.com/}, one of the leading payment platforms, provides the users with convenient payment service as well as the security assurance. The data scientists take efforts to build up models and confront the frauds.

With the develop of Alipay, potential challenges arise\cite{zhang2019distributed}. In the fraud detection problem, it is intractable to accumulate enough training data for the models. Especially, the positive samples are collected by case reports from the victims, which indicate the invocatable losses have come into being. On one hand, the data scientists need labels to train the fraud detection models, on the other hand, they are eager to protect the users from frauds. Therefore making full use of existing data is crucial. While providing online payment service to the merchants, the risk management mechanism is set up for each individual merchant, but some small merchants even do not have enough samples to train a usable model. The merchants would like to share the anti-fraud expertise across to form a stronger guard. Another practical scenario is how to quickly build up a risk management system for a new marketplace. In face of the cold start problem, it's a natural choice to refer to the data under similar scenes. 

Transfer learning is born to solve these problems. It helps to transfer the information from the source domain, with abundant data, to the exhausted target domain. In this way, the source domain data are reused wisely and the data required to build up a practical model on the target domain is reduced.

Nevertheless, there are implicit restrictions in previous cases. Due to data privacy protection regulations, e.g. General Data Protection Regulation(GDPR)\footnote{https://eugdpr.org/}, cross-boarder data-sharing is under increasingly strict supervision. The merchants are also unwilling to leak any customer infomation to their competitors. So the model-based transfer learning is the only solution, instead of feature-based and instance-based ones. What's more, the model-based methods also work well when the storage capacity and transmission throughout on the target domain are limited.

Most of the model-based transfer learning algorithms are based on the neural network, which are good at dealing with the image, audio and natural language data. However, the fraud detection features are well-defined and the feature number is relatively small. Neural network is not the optimal choice here, in that the parameters are hard to tune and the model is hard to interpretate. Gradient Boosting Desition Tree(GBDT) is an ensemble model with appealing characteristics, e.g. the ability to handle nonlinearity and not requiring the tedious feature pre-processing. It is a popular option for the machine learning tasks and the following variants optimize the original algorithm in both performance and speed. Many excellent implementations, such as XGBoost\footnote{https://github.com/dmlc/xgboost} and LightGBM\footnote{https://github.com/microsoft/LightGBM}, are released and widely adopted.

In this paper, we focus on the above mentioned real-world fraud dection problem and propose an adapted tree boosting workflow for transfer learning. We implement the algorithm on top of XGBoost and get a competitive result in the experiment. The rest of the paper is organized as follows. Section \ref{secrel} provides a brief review of related work on transfer learning. After that, section \ref{secpre} formalizes the basic transfer learning settings as the preliminary. And the key insights and necessary computation details of GBDT and XGBoost are also introduced. In section \ref{secmech}, we first analyze the data and describe the main framework. Then intuitions and related operations to revise a source tree model are explained in detail. Section \ref{secexp} contains our experiments and discussions, followed by the conclusion in section \ref{seccon}.

\section{Related Work}\label{secrel}
Transfer learning approaches are designed to transfer the helpful information from the source task to a similar target task and the information is exploited to promote the performance on the target task. Four classic paradigms are proposed based on what to transfer\cite{pan2009survey}. Instance-based methods reuse the souce domain data by instance weighting and importance sampling\cite{khan2016adapting,zadrozny2004learning,cortes2008sample,tan2017distant}. These methods work well when the data distributions in the source domain and the target domain are sufficiently approximate. Feature-based methods transform the features from both domains to a proper feature space to reach the goal\cite{liu2011cross,zheng2008transferring,hu2011transfer,long2014transfer,duan2012domain,blitzer2006domain}. They try to find a feature representation that minimizes the difference between the domains. Relation-based methods help the relational domains and focus on the the relation mapping between the domains\cite{mihalkova2007mapping,mihalkova2008transfer,davis2009deep}.
Model-based methods assume that the source model can contribute to the target model and they share parameters between the domains\cite{ge2013oms,zhao2011cross,pan2008transferring}. Common approaches including adapting models with the biased regularizers\cite{Kienzle2006Personalized,rodner2008learning,tommasi2010safety,rodner2011learning}, ensemble strategies\cite{rettinger2006boosting,luo2008transfer,ruckert2008kernel,baxter2000model} and utilizing source model as priors\cite{pratt1991direct,thrun1994learning,eaton2008modeling}.

Model-based methods, which we adopt in this work, stand out in the situation where the source domain is referential but the samples are not available to the target domain. There are attempts for different models in this field. Zhao et al. proposes TransEMDT that integrates a decision tree and the k-means clustering algorithm to build a personalized activity recognition model\cite{zhao2011cross}. A decision tree is trained offline and unlabeled samples from a new user are then used to adapt the model. Another similar work is based on extreme learning machine\cite{Deng2014Cross}. SVM is a popular base model. ASVM(Adapted SVM)\cite{Yang2007Cross} and its extension\cite{Duan2009Domain} train a Gaussian kernel SVM as source model and regularize it with target samples. Many recent researches integrate neural network into transfer learning in order to learn the feature representation or share the model parameters. The adaptation layers and domain related losses are introduced to design customized neural network architectures\cite{long2015learning,long2016deep,long2017deep,Tzeng2015Simultaneous}. Though the deep neural networks get fruitful results in natural language, audio  and image data, tree models are more competitive when the features are well-constructed in limited feature dimensions. Considering the tree models are easy to train and interpretate, we select tree model as the base learner.

The tree-based transfer learning is widely explored for streaming data mining. The methods are designed to process massive, high-speed streams and modify the decision tree with new samples. In \cite{domingos2000mining,hulten2001mining, jin2003efficient}, the decision tree is constructed on the streaming data and proved to be asymptotically nearly identical to the tree of a conventional learner. The algorithm is optimized in both time(worst-case proportional to the number of attributes) and space(proportional to the size of the tree and associated sufficient statistics). N{\'u}{\~n}ez et al. proposes an incremental decision tree that automatically adjusts its
internal parameters\cite{nunez2007learning}. Target concepts in streaming data change over time and a time window with adaptive size is introduced to solve the problem. A local performance measure is calculated and when the performance on a leaf decreases, the size of its local window is reduced. These methods process the changing batch data rather than a constant target dataset.

The model-based transfer learning framework raised by Segev et al.\cite{segev2016learn} is similar to this work, in that it also consists of several revise strategies operating on the transfered model. Two algorithms are applied to revise the tree structures of the random forest trained in the source domain. The SER(structure expansion/reduction) algorithm first expands the leaf node by utilizing the samples falling into it. Then the reduction operates on the intermediate nodes in a bottom-up order and generalizes the rules by pruning. In order to prune the tree, two types of errors are recorded: subtree error on a node is the empirical error of the its descendants and leaf error is the error when the node is  regarded as a leaf node. If the leaf error is smaller than subtree error, the subtree dominated by this node is then pruned into a leaf node. The STRUT(structure transfer) algorithm is designed from a new perspective. The scales of features are observed to be quite different between the domains, so the associated decision thresholds should be modified. A final MIX method is employed to combine the two result forests generated by SER and STRUT as a voting ensemble. This framework is different from ours because of the different essence of base models. Random forest is a typical bagging model and a bunch of decision trees vote together to get the prediction. Boosting tree models outperforms random forest and could be utilized differently in transfer learning.

A well-known transfer learning algorithm based on boosting tree models is the TrAdaBoost\cite{dai2007boosting}. It inherits the key idea of AdaBoost\cite{freund1997decision} and adjusts the weights of training samples. The wrongly predicted samples in source domain are regard as misleading data and their effects are weaken in the following iterations. Besides the normal AdaBoost adopted in target domain, TrAdaBoost defines a multiplier to decrease the weights of misclassified source samples. After several iterations, the most helpful diff-distribution data are picked out as supplement for the same-distribution data. AdaBoost is equivalent to GBDT with exponential loss function. We choose the more generic GBDT as base model and carry out the work under a model-based transfer learning paradigm.

\section{Preliminary}\label{secpre}
Our method serves the scarce dataset as a transfer learning framework in order to provide a usable machine learning model. We take the gradient boosting tree algorithm as the basis of this framework. In this section, we briefly introduce some necessary notations and concepts used in the algorithm. 

\subsection{Transfer Learning Settings}
The key definition {\em domain} in transfer learning consists of 3 factors: $ \mathcal{D} = (\mathcal{X}, \mathcal{Y}, P)$, where $\mathcal{X}$ and $\mathcal{Y}$ stands for the d-dimension feature space and label space respectively. $P$ represents the probability distribution of the data, which is a virtual oracle without the explicit representation. There are 2 basic domains in transfer learning. Source domain contains abundant labeled data and target domain, on the contrary, provides scarce information. The transfer learning tasks attempt to transfer the knowledge included in source domain data to the target domain.

We introduce $s$ and $t$ as subscript to represent the source domain and target domain as $\mathcal{D}_{s}$ and $\mathcal{D}_{t}$. There are various data hypotheses for different transfer learning algorithms. Obeying our real-world scenario, we follow the restriction that the source domain and target domain share the same feature space $\mathcal{X}$ and label space $\mathcal{Y}$, i.e. we have $\mathcal{X}_{s} = \mathcal{X}_{t}=\mathcal{X}$ and $\mathcal{Y}_{s} = \mathcal{Y}_{t}=\mathcal{Y}$.

Supposing that there are $n_{s}$ instances in source domain and $n_{t}$ instances in target domain. $X_{s}$ is employed to denote the feature matrix in source domain where $X_{s}$ is an $n_{s} \times d$ matrix and $X_{s} = \{x_{i}\}_{i=1}^{n_{s}}$. Dataset $D_{s}$ consists of feature matrixes and related label vectors $D_{s}=\{X_{s}, Y_{s}\}$. Similarly, we have the feature matrix in target domain $X_{t} = \{x_{j}\}_{j=1}^{n_{t}}$ and $D_{t}=\{X_{t}, Y_{t}\}$.

The marginal distributions of features are always different between $\mathcal{D}_{s}$ and $\mathcal{D}_{t}$ in transfer learning, $P_{s}(x_{s}) \ne P_{t}(x_{t})$. Further, the conditional probability distribution of labels are different in our settings, $Q_{s}(y_{s}|x_{s}) \ne Q_{t}(y_{t}|x_{t})$. Further, We apply the model-based transfer method. Our final goal is to get a classifier for target domain $F_{t}: x_{t} \mapsto y_{t}$. To reach the goal, a model is first built in the source domain $F_{s}: x_{s} \mapsto y_{s}$. We then take $F_{s}$ as the knowledge transferred from the source domain and make some adaptations to guide the model building in target domain.

\subsection{Gradient Boosting Decision Tree}
The machine learning tasks are generally solved as function estimation problems. A proper loss function $L(y, F(x))$ is defined to measure the difference between the labels and the predictions. Then the optimization goal of function estimation is minimizing the expected value over the joint distribution of all training samples.
\begin{equation}
F^{*} = \mathop{\arg\min}\limits_{F}E_{x, y}L(y, F(x)) \label{formula7}
\end{equation}

GBDT carries out the estimation in a nonparametric way and applies numerical optimization in function space\cite{friedman2001greedy}. The prediction $F(x)$ at each point $x$ is regarded as a parameter and the minimizing objective is deduced as follows directly with respect to $F(x)$.
\begin{equation}
\begin{aligned}
\phi(F(x)) &= \mathop{\arg\min}\limits_{F(x)}E_{x, y}L(y, F(x)) \\
&= \mathop{\arg\min}\limits_{F(x)}E_{x}[E_{y}(L(y, F(x))|x)] \\
&= {\mathop{\arg\min}\limits_{F(x)}E_{y}[L(y, F(x))|x]}
\end{aligned}
\end{equation}

The optimal $F^{*}$ is solved as a numerical optimization. The gradient descent method is applied and $F^{*}$ is denoted in an additive form. Where $f_{0}$ is the initial value and followed by $M$ boosting values$\{f_{m}\}_{1}^{M}.$ At each step, the descent direction is determined by the gradient $g_{m}$, which is in infinite-dimensional space. Then the step size $\rho_{m}$ is computed by minimizing the objective in the negative gradient direction.
\begin{eqnarray}
F^{*} &=& \sum\limits_{m=0}^{M}f_{m} \\
g_{m} &=& \{g_{im}\} = \{\frac{\partial{E_{y}[L(y, F(x))|x_{i}]}}{\partial{F}}|_{F=F_{m-1}}\} \\
\rho_{m} &=& \mathop{\arg\min}\limits_{\rho}E_{x, y}L(y, F_{m-1}-\rho g_{m})\label{formula1} \\
f_{m} &=& -\rho_{m}g_{m}
\end{eqnarray}

However, real-world datasets contain finite training samples and $F^{*}$ cannot be estimated at each $x_{i}$. In order to smooth the estimation among the samples and generalize to the data beyond training set, a function $h$ with parameter $\alpha$ is introduced to fit the gradients at each step.
\begin{equation}
g_{im} = g_{m}(x_{i}) \simeq h(x_i;\alpha_{m})
\end{equation}

The optimization turns into finding the best $h_{m}=\{h(x_{i};\alpha_{m})\}_{1}^{N}$ that most parallel to $-g_{m} \in R^{N}$, where $N$ is the number of samples. The parameter $\alpha_{m}$ is obtained by solving that:
\begin{equation}
\alpha_{m} = \mathop{\arg\min}\limits_{\alpha, \beta}\sum\limits_{i=1}^{N}[-g_{m}(x_{i})-\beta h(x_i;\alpha)]^{2} \label{formula2}
\end{equation}

With the direction $h(x_{i};\alpha_{m})$ and finite dataset, the step size in (\ref{formula1}) and optimal estimation in (\ref{formula7}) is updated:
\begin{eqnarray}
\rho_{m} &=& \mathop{\arg\min}\limits_{\rho}\sum\limits_{i=1}^{N}L(y_{i}, F_{m-1}(x_{i})+\rho h(x_i;\alpha_{m})) \label{formula4} \\
F_{m}(x)&=& F_{m-1}(x) + \rho_{m}h(x_i;\alpha_{m}) \label{formula3}
\end{eqnarray}

Function $h$ is also called the base learner.  $F^{*}$ is searched in the function space which is restricted by $h(x_i;\alpha)$. GBDT chooses decision tree as the base learner. A tree model is parameterized by each leaf node region $R_{j}$ and its weight $b_{j}$, then a tree model with $J$ leaf nodes is represented as
\begin{eqnarray}
h(x;\{b_{j}, R_{j}\}_{j=1}^{J}) = \sum\limits_{j=1}^{J}b_{j}\mathbb I(x\in R_{j})
\end{eqnarray}

where the function $\mathbb I$ indicates whether a sample $x$ falls into the leaf node $j$.

By solving (\ref{formula2}), the leaf weight is the weighted average of the latest gradients. 
\begin{equation}
b_{jm} = \frac{\sum\limits_{x_{i}\in R_{jm}} w_{i}g_{im}}{\sum\limits_{x_{i}\in R_{jm}} w_{i}} = \bar{g}_{m}
\end{equation}

With this weight estimation, the split operation can be evaluated by computing the gain of objective in (\ref{formula2}). By traversing the features and values, the tree is construct node by node and the structure is finally determined. The line search multiplier $\rho_{m}$ can be merged in $b_{jm}$ and (\ref{formula3}) is expressed as:

\begin{equation}
\begin{aligned}
F_{m}(x) &= F_{m-1}(x) + \rho_{m} \sum\limits_{j=1}^{J} b_{jm}\mathbb I(x\in R_{j}) \\
&=F_{m-1}(x) +  \sum\limits_{j=1}^{J}\gamma_{jm}\mathbb I(x\in R_{j})
\end{aligned}
\end{equation}

According to (\ref{formula4}) the leaf weight is updated as:
\begin{equation}
\gamma_{jm}=\mathop{\arg\min}\limits_{\gamma}\sum\limits_{x_{i}\in R_{j}}L(y_{i}, F_{m-1}(x_{i})+\gamma)
\end{equation}

\subsection{XGBoost}
The gradient decent in GBDT is a first-order optimization algorithm. The variants, such as XGBoost\cite{chen2016xgboost} and Psmart\cite{zhou2017psmart}, add regularizer $\varOmega$ to the objective and the objective at $m$ iteration is
\begin{equation}
O^{(m)} =\sum\limits_{i=1}^{N}L(y_{i}, F_{m-1}(x_{i})+f_{m}(x_{i}))+\varOmega(f_{m})
\end{equation}

The objective takes the second-order taylor expansion approximation of objective and regularizes the number of leaf nodes $J$ and leaf weights $\gamma$. It can be regrouped by leaf:
\begin{equation}
\begin{aligned}
O^{(m)}&\simeq \sum\limits_{i=1}^{N}[g_{im}f_{m}(x_{i})+\frac{1}{2}h_{im}f_{m}^{2}(x_{i})]+(\eta J+\frac{1}{2}\lambda \sum\limits_{j=1}^{J}\gamma_{jm}) \\
&=\sum\limits_{j=1}^{J}[(\sum\limits_{x_{i}\in R_{jm}}g_{im})\gamma_{jm}+\frac{1}{2}(\sum\limits_{x_{i}\in R_{jm}}h_{im}+\lambda)\gamma_{jm}^{2}]+\eta J \\
&=\sum\limits_{j=1}^{J}[G_{jm}\gamma_{jm}+\frac{1}{2}(H_{jm}+\lambda)\gamma_{jm}^{2}]+\eta J
\end{aligned}
\end{equation}
where $G_{jm}$ and $H_{jm}$ is the first-order and second-order derivative summation on leaf $j$.

Then the optimal weight and related objective are:
\begin{eqnarray}
\gamma_{jm} &=& -\frac{G_{jm}}{H_{jm}+\lambda} \label{formula5} \\
O^{(m)} &=& -\frac{1}{2}\sum\limits_{j=1}^{J}\frac{G_{jm}^{2}}{H_{jm}+\lambda}+\eta J \label{formula6}
\end{eqnarray}
Formula (\ref{formula6}) is employed to evaluate the split and construct the tree. After that, the leaf weight is calculated by (\ref{formula5}).

LogLoss is a common loss function for classification problem and it is calculated by label $y$ and output probability $\hat y$, where $\hat y = \frac{1}{1+e^{F_{m}}}$.
\begin{equation}
L(y, \hat y) = -yln(\hat y)-(1-y)ln(1-\hat y)
\end{equation}

With this definition, the first-order derivative $g_{m}$ and second-order derivative $h_{m}$ used in the following revise algorithms can be formalized:
\begin{eqnarray}
\begin{aligned}
g_{m}&=\frac{\partial{L}}{\partial{F_{m}}}\\
&=y\cdot \frac{1}{1+e^{-F_{m}}}\cdot e^{-F_{m}}\cdot(-1)+(1-y)\cdot\frac{1}{1+e^{F_{m}}}\cdot e^{F_{m}}\\
&=-y\cdot(1-1+e^{-F_{m}})+(1-y)\cdot\frac{1}{1+e^{-F_{m}}} \\
&=-y+\frac{1}{1+e^{F_{m}}} \\
&=\hat y-y \\
h_{m}&=\frac{\partial{L}}{\partial^{2}{F_{m}}}=(-1)\frac{e^{-F_{m}\cdot(-1)}}{(1+e^{-F_{m}})^{2}}=(1-\hat y)\cdot \hat y
\end{aligned}
\end{eqnarray}
\section{Mechanism}\label{secmech}
To face the fact that there are no sufficient data to obtain practical models under new scenes. Data, especially the labels, are time-consuming and costly to collect. So it's natural to seek for some help from the accumulated data under similar scenes, namely source domain data in transfer learning. In this section, we start from our scenario and design a transfer learning framework to utilize the treasurable data based on the XGBoost.

\subsection{Data Analysis}\label{secdata}
Both of the data in our source domain and target domain are drawn from the oversea e-commerce transactions, but in different marketplaces. Here, the features are defined based on the fraud detection expertise and describe the characteristics distinguishing the samples with different labels. Obviously, these characteristics can be shared across different fraud scenes in a degree.

However, the data distributions can be various in several ways. Different fraud groups take charge of different regions and different fraudsters may follow different fraud tricks. Another point is that their behaviors are completely related to the local risk management strategies. The region-wise strategies are formulated differently, so the fraudsters have to adjust their actions along with the strategies even for the same vulnerability.

Several drift modes show up in Fig.\ref{figdist}, which compares the feature distributions between domains :
\begin{figure}[htbp]
 \centering
 \includegraphics[width=0.45\textwidth]{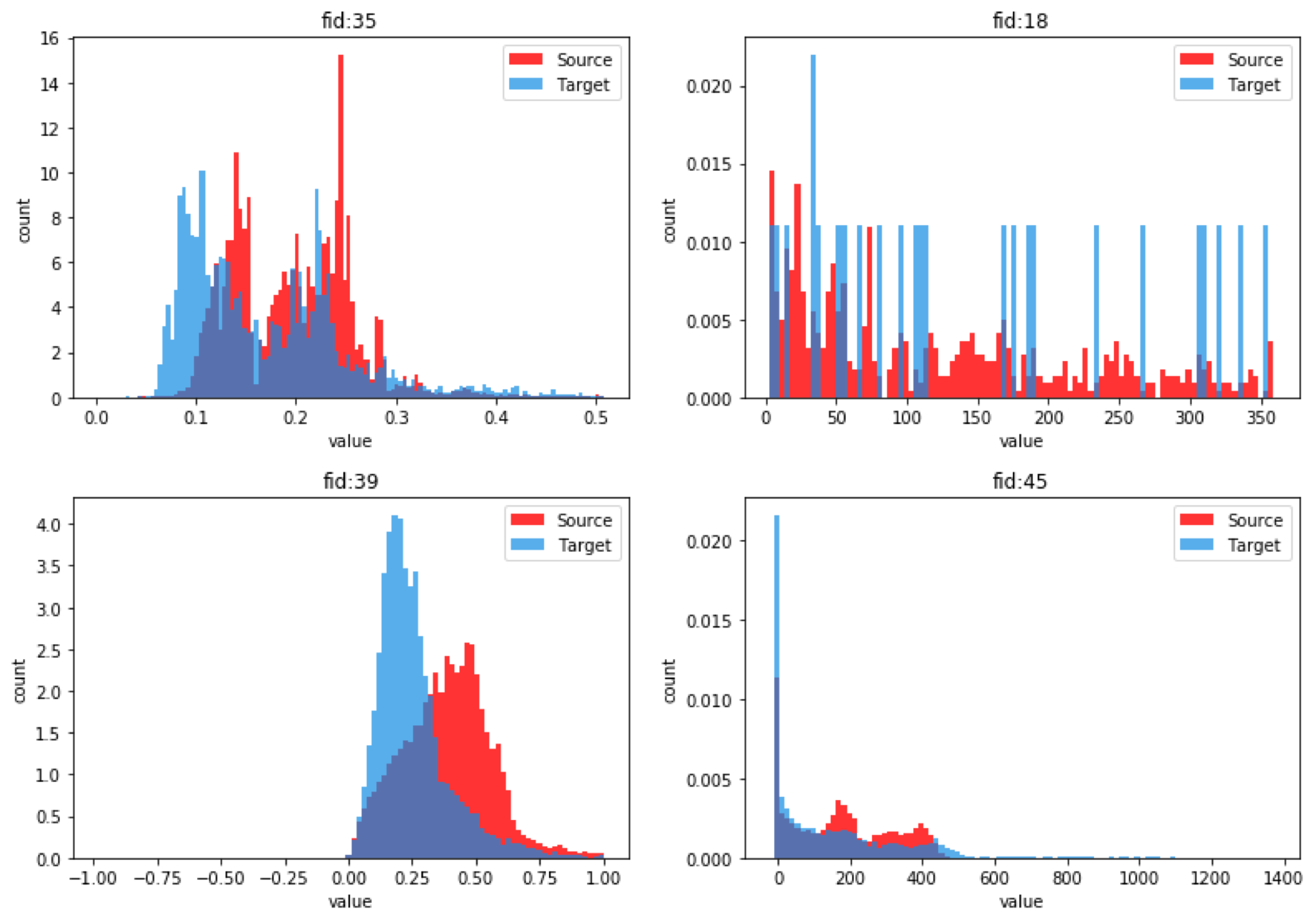}
 \caption{Feature distribution between domains}\label{figdist}
\end{figure}

\begin{itemize}
\item Only the range of feature value is changed, while the shape of distribution remains similar(fid35)
\item The shape of feature distribution changes(fid39 \& fid45)
\item The feature lose efficacy and is not so active under the new scene(fid18)
\end{itemize}

Additionally, we can also observe the predictive power change of the features from the IV(Information Value) metric. As shown in TABLE \ref{tabiv}, the top IV features on different domains vary a lot and one third of the top IV features on the source domain become less predicative(marked as red). Besides the relative rank, the overall value scale of IV and the IV of the same feature are also different. This reveals the relation change between the features and the labels.

\begin{table}[htbp]
  \centering
  \caption{The IV Comparison of Top 20 Features on Source Domain}
    \begin{tabular}{l|ll|ll|l}
    \toprule
    \textbf{feat\_id} & \textbf{IV$_{s}$} & \textbf{rank$_{s}$} & \textbf{IV$_{t}$} & \textbf{rank$_{t}$} & \textbf{rank\_diff} \\
    \midrule
    43    & 1.412 & 1     & 0.9   & 3     & 2 \\
    44    & 1.412 & 2     & 0.9   & 4     & 2 \\
    5     & 1.268 & 3     & 0.374 & 26    & \textcolor[rgb]{ 1,  0,  0}{23} \\
    17    & 1.01  & 4     & 0.629 & 7     & 3 \\
    34    & 0.749 & 5     & 0.5   & 11    & 6 \\
    37    & 0.65  & 6     & 0.283 & 31    & \textcolor[rgb]{ 1,  0,  0}{25} \\
    35    & 0.634 & 7     & 0.488 & 13    & 6 \\
    36    & 0.629 & 8     & 0.341 & 27    & \textcolor[rgb]{ 1,  0,  0}{19} \\
    38    & 0.507 & 9     & 0.565 & 9     & 0 \\
    6     & 0.369 & 10    & 0.384 & 23    & \textcolor[rgb]{ 1,  0,  0}{13} \\
    45    & 0.337 & 11    & 1.024 & 1     & 10 \\
    47    & 0.337 & 12    & 1.024 & 2     & 10 \\
    46    & 0.264 & 13    & 0.797 & 5     & 8 \\
    31    & 0.236 & 14    & 0.449 & 15    & 1 \\
    8     & 0.228 & 15    & 0.232 & 34    & \textcolor[rgb]{ 1,  0,  0}{19} \\
    28    & 0.224 & 16    & 0.423 & 17    & 1 \\
    1     & 0.219 & 17    & 0.38  & 24    & 7 \\
    30    & 0.208 & 18    & 0.428 & 16    & 2 \\
    41    & 0.206 & 19    & 0.596 & 8     & 11 \\
    15    & 0.202 & 20    & 0.271 & 32    & \textcolor[rgb]{ 1,  0,  0}{12} \\
    \bottomrule
    \end{tabular}%
  \label{tabiv}%
\end{table}%

\begin{figure}[htbp]
 \centering
 \includegraphics[width=0.45\textwidth]{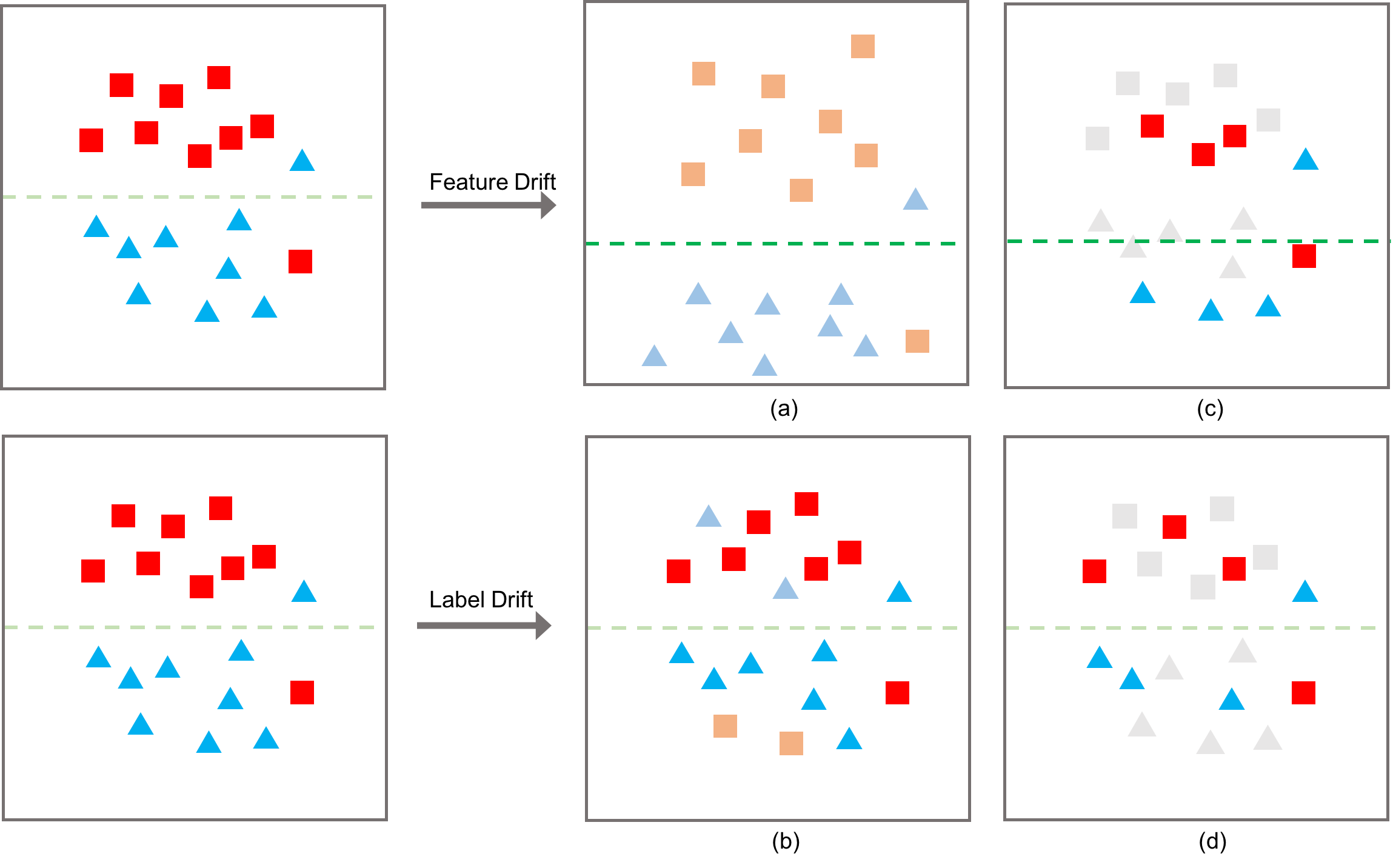}
 \caption{Feature and label drift between domains}\label{figdrift}
\end{figure}

In fact, both of the labels and features can be shifty from our analysis of the data difference between the domains. Fig.\ref{figdrift} visualizes the typical drifting cases. The fraudsters change their operandi over time and are active at different regions. So the feature drifting in Fig.\ref{figdrift}-a often takes place from one scene to another. Meanwhile, the label drifting in figure Fig.\ref{figdrift}-b may result from the rule change in the definition of ``bad'' under different scenes. Sometimes, even our datasets are subsamples from the same oracle, a drift can be brought out with the limited sample counts as in Fig.\ref{figdrift}-c and Fig.\ref{figdrift}-d.

Though there are distribution changes and relation changes among the domains, the fraudsters do share similar patterns and attributes, which is in harmony with the basic supposition of transfer learning. In particular, the data experts would like to refer to mature scenes and make the defense plans under some new scenes with rare data or labels. The experts manually extract the similarities and this process works out as a good initialization of cold starting, which proves the feasibility of utilizing transfer learning methods to help with the data deficiency.

\subsection{Model-based Transfer}\label{subsecmodbase}
In our case, we have to protect the privacy of the original data and the data are in separated projects. So we turn to model-based transfer learning, a light-weight transfer from the source domain to the target domain.

GBDT is a common choice in machine learning tasks. Besides the high performance and efficiency, GBDT and its variants also provides the model interpretability\cite{fang2018unpack} and the easiness of parameter tuning. The most direct transfer is first train a model on the source dataset. Then the model is transferred to target domain and further trained based on the target dataset. The final model is a fine-tuned version, which inherits the informative patterns in source domain and fits in with the target dataset. Sharing the model directly is a good solution if the target dataset stay the same as the source dataset, but it may be fruitless due to the data drifting in previous analysis. Considering the above reasons, we choose XGBoost as the basic learner of this transfer learning framework and make adaptations to the source trees before continuing training.

Algorithm \ref{alg:xgbtl} shows the main workflow of our training process.  Given a training dataset $\mathcal{D}$, the wrapper function $TrainXGB$ trains $T$ trees with $L$ levels based on a base model $M$. It calls the XGBoost API and takes advantage of this highly optimised toolkit. There are two strategies listed to conduct the training process, namely OneRound and MultiRound respectively. The function names indicate the number of times needed for transferring models between the domains. 
\begin{itemize}
\item OneRound: train a batch of trees on the source domain and transfer them to the target domain in bulk. Then on the target side, the trees are revised one by one and become the base model for further training.
\item MultiRound: iteratively train on the source domain and revises on the target domain to grow each source tree. Instead of the batch process,  multiple model exchanges take place during the base model training. Then continue training with only target dataset.
\end{itemize}

The main difference is how to train the source model. The OneRound strategy captures more patterns in the source domain and maintains a lower transmission cost. The MultiRound strategy fits the target domain better and considers the statistics change after tree revising. Which strategy to apply is a hyper parameter to be decided based on the datasets.

\begin{algorithm}[htb] 
  \begin{algorithmic}[1]  
  \caption{Transfer Learning XGB}
   \label{alg:xgbtl}  
   \Function {OneRound}{$\mathcal{D}_{s}, \mathcal{D}_{t}, T_{s}, T_{t}, L_{s}, L_{t}$}
   \State $M_{s} = TrainXGB(\mathcal{D}_{s}, T_{s},  L_{s},M=None)$
   \State // transfer $M_{s}$ to the target domain
   \State $M_{s}' = [\ ]$
    \For{$i=1,2,...,T_{s}$}
    \State $ tree = ReviseOneTree(M_{s}', M_{s}[i-1], \mathcal{D}_{t}) $
    \State $M_{s}'.append(tree)$
    \EndFor 
    \State $M_{t} = TrainXGB(\mathcal{D}_{t}, T_{t},  L_{t}, M=M_{s}')$
    \State \Return $M_{t}$
     \EndFunction    
   \Function {MultiRound}{$\mathcal{D}_{s}, \mathcal{D}_{t}, T_{s}, T_{t}, L_{s}, L_{t}$}
    \State $M_{s} = [\ ]$
    \State $M_{s}' = [\ ]$
    \For{$i=1,2,...,T_{s}$}
    \State $ M_{s} = TrainXGB(\mathcal{D}_{s}, 1,  L_{s},M=M_{s}) $
    \State // transfer $M_{s}[i-1]$ to the target domain
    \State $ tree = ReviseOneTree(M_{s}', M_{s}[i-1], \mathcal{D}_{t}) $
    \State $M_{s}'.append(tree)$
    \State // transfer $M_{s}'[i-1]$ to the source domain
    \State $M_{s}[i-1] =M_{s}'[i-1]$
    \EndFor 
    \State $M_{t} = TrainXGB(\mathcal{D}_{t}, T_{t},  L_{t}, M=M_{s}')$
    \State \Return $M_{t}$
    \EndFunction
   \end{algorithmic}  
\end{algorithm}

\subsection{Revise Strategies}\label{subsecrevise}
In Algorithm \ref{alg:xgbtl}, the function $ReviseOneTree$ takes current revised model $M_{s}^{'}$ and next source tree $M_{s}[i-1]$ as input, then conduct the revise strategies based on the target dataset $\mathcal{D}_{t}$. In this section, we discuss some practical strategies, inspired by the data drifting analysis, and related implementation details.

\subsubsection{Re-split}
As shown in Fig.\ref{figdist} and Fig.\ref{figdrift}-a, the most common feature drifts are scale change and shape change, which fade the branches of the source model. From our observation, the original split values usually result in an unbalanced sample division and the source patterns then become useless. It's obvious that the value is no longer the best split point under the new data distribution. To maintain the efficacy, recomputation of split value is necessary. 

The original split features, which construct the backbone of the valuable patterns, are reserved while re-spliting the target dataset. Reviewing the computation of split gain in XGBoost in Formula (\ref{formula6}), the first-order and second-order derivatives rely on current predict values and labels. Firstly, the $g$ and $h$ of each sample could be computed, with which we get the split gain for every possible feature values. Then we traverse the whole tree in a top-down order and revise all the split values.

When the data drifting belongs to the scale change but the labels are rare, a simpler method also works out fine. We can first compute the fractile of split value in source dataset and get the same fractile in target dataset as new split value.

\subsubsection{Re-weight}
After getting the tree structure, XGBoost assigns a score to each leaf node with Formula (\ref{formula5}). The leaf score is determined by the sample labels fallen into the it. To deal with the label drift problem in Fig.\ref{figdrift}-b and eliminate the effects of the distribution change, we have to re-weight the leaves. It's easy to embed the score computation in the steps of re-split.

\subsubsection{Rare Branch}
A common case, when we apply the revise strategies, is that some of the branches only contain a small amount of samples. We lose the statistical significance to re-split and may overfit on the these samples. In this way, the revise neither make better use of the source dataset nor result in a reasonable final model. So the sample count on every subtree should be recorded and taken into consideration while executing the strategy. 

For the branches with no sufficient samples, two new operations are needed: prune and discount. Prune means that we set up a least sample threshold and drop a rare branch directly. By doing so, a intermediate node becomes a leaf node. This method can cut off the source patterns that is invalid. However, for the cases in Fig.\ref{figdrift}-c and Fig.\ref{figdrift}-d, rare branch is caused by the limited dataset and the related pattern may work well on future samples. Because we are not confident of the rare branches, a discount factor is multiplied to the original weight, as a hyper parameter to be determined. This factor reflects the confidence that a source pattern will take effect on the target domain. We tempted to maintain the original branch information, only to get a bad result. The reason is that the original scores are confused with the revised scores, but they are not comparable. With the factor, we weaken the source scores.

TABLE \ref{tabiv} shows that some features may lose efficacy and fail to be a good split feature. When we traverse the samples and compute the new split gain, no positive result could be found and the feature cannot bring information gain by dividing the samples. So we have to treat it just as the rare branches and apply the prune/discount strategies.

\subsubsection{Implementation}
The basic train and predict process are provided by the XGBoost toolkit. To implement the revise strategies, a data structure is in need to restore the necessary tree structure and the intermediate split statistics. A parser is designed to initialize the data structure from the dumped XGBoost model file.

In order to continue training an XGBoost model, the data structure should also support reverse model transformation. However, no API is released in XGBoost to load a dumped model. In this case, we employ the Treelite\footnote{https://github.com/dmlc/treelite} to solve this problem. Treelite is compatible with the most popular GBDT toolkit, e.g. XGBoost, LightGBM and Scikit-learn\footnote{https://scikit-learn.org/stable/}. It allows us to define a Treelite model and generate a mocked binary model, which XGBoost is able to load.



\section{Experiment}\label{secexp}
The experimental dataset contains the card(debit card or credit card) transaction samples from one of the oversea e-commerce platform partners of Alibaba, which utilizes the risk management system of Alipay to detect the real-time transaction frauds, e.g. card-stolen cases. There are nearly 5 million successful transaction samples from 5 contries in total, which consists of basic transaction information, fraud labels and 48 well-designed features. 

The features are designed to describe different risk factors. For example, whether the shipping information and device of the payment are reliable, the consistency of the user behaviors and the buyer-seller relation, etc. 
The fraud labels are collected in the following 3 ways:
\begin{itemize}
\item Chargeback from the card issuer banks: the card issuer receives claims on unauthorized charges from the card holders and report related transaction frauds to the merchants
\item Third-party data: the fraud data, gathered from banks worldwide, from the venders work as a good supplement
\item Label propagation: the device and card information are also utilized to mark similar transactions
\end{itemize}

\subsection{Experiment setup}
Because some scenes are new or the labels need time to be fed back, the bad rate in different contries are quite different(TABLE \ref{tabstat}). We take the mature scene with highest bad rate as the source domain and the rest scenes as the target domain. In order to simulate the common case of real-world applications, we take 30\% of the target domain as training data. In this way, the training data in target domain cannot build a robust model and the evaluation is more reliable with more positive labels. To better understand the data, We run a basic XGBoost model in the dataset in advance. Because of the low bad rate, we apply the negative sampling method to get a better performance. After this, the bad rate of the training data on both domains is about 3\%.

\begin{table}[htbp]
  \centering
  \caption{Experimental Dataset Statistics}
    \begin{tabular}{lllll}
    \toprule
    \textbf{scene\_id} & \textbf{neg\#} & \textbf{pos\#} & \textbf{bad\_rate} & \textbf{domain} \\
    \midrule
    3     & \textcolor[rgb]{ 0,  .439,  .753}{1227023} & \textcolor[rgb]{ 0,  .439,  .753}{1872} & \textcolor[rgb]{ 0,  .439,  .753}{0.152\%} & \textcolor[rgb]{ 0,  .439,  .753}{Source} \\
    \midrule
    \multicolumn{1}{p{5.25em}}{1} & 777364 & 368   & 0.047\% & \textcolor[rgb]{ .816,  .808,  .808}{} \\
    \multicolumn{1}{p{5.25em}}{2} & 1285753 & 74    & 0.006\% & \textcolor[rgb]{ .816,  .808,  .808}{} \\
    \multicolumn{1}{p{5.25em}}{4} & 371938 & 96    & 0.026\% & \textcolor[rgb]{ .816,  .808,  .808}{} \\
    \multicolumn{1}{p{5.25em}}{5} & 1299483 & 464   & 0.036\% & \textcolor[rgb]{ .816,  .808,  .808}{} \\
    \midrule
    \textcolor[rgb]{ .267,  .447,  .769}{} & \textcolor[rgb]{ 1,  0,  0}{3734538} & \textcolor[rgb]{ 1,  0,  0}{1002} & \textcolor[rgb]{ 1,  0,  0}{0.027\%} & \textcolor[rgb]{ 1,  0,  0}{Target} \\
    \bottomrule
    \end{tabular}%
  \label{tabstat}%
\end{table}%

Here list the main metrics we care about:
\begin{itemize}
\item AUC: basic discriminability measurement for classification problem, area plotted with TPR against the FPR
\item Top-Recall: recall of the samples whose scores rank on the top, in practice, the 1\textpertenthousand\ recall is in use 
\end{itemize}
While evaluating the dataset with the class imbalance problem, AUC will be high and the gap between models will be minor. In this case, Top-Recall is a better measurement. What's more, it plays a import role in fraud detection, because the follow-up actions are taken on the most suspicious users. To provide a better user experience, it's necessary to ensure the model performance at the top.

From the description in section \ref{subsecrevise}, the algorithm involves many parameters. One part of the parameters are related to XGBoost, we focus on the most important two: the tree number and the tree depth. And these XGBoost parameters on source and target domains should be tuned respectively. Other parameters, like learning rate and sample rate, we fix them up globally and initialize them with the aforementioned basic XGBoost model.
Another part of the parameters control the revise process, e.g. whether execute the re-split, re-weight, prune and the proper discount of bare branches. 

The Baseline Model 1(BM1) trains model with training set only on the target dataset. The Baseline Model 2(BM2) transfers the source model without any revise. We can conclude the gain of transfer learning and the effects of tree revision by comparing the model performance with BM1 and BM2 respectively.

\subsection{Revise Analysis}
We recorded the statistics on each tree node during training the XGBoost model on source domain and compared them with the statistics in the revising process. Here, the first tree serves as an example, we'd like to analysis our observations on the revise strategies.

\subsubsection{Re-split}
The intermediate nodes with more than 1 sample on the target domain are listed in TABLE \ref{tabresplit}. We print out the split value, split gain and sample number on both domains. The score is calculated just as the leaf node and it reflects the label distribution in the node.
\begin{itemize}
\item There are nodes(marked as red) that cannot get a positive split gain. The reasons could be lacking in samples(node 12) or pattern difference between domains(node 19). It's pointless to split on and the following path should be skipped.  
\item Observing the nodes with statistical significance (containing more than 3000 samples), the new split values are quite different from the old ones. The original split values tend to get imbalanced samples divisions with the drifted features.
\item We rank the nodes with positive re-split gains by the score difference. As we can see, the node with low score difference(marked as green) tends to be statistically significant. It indicates our datasets have minor label drift. What's more, large score differences(marked as blue) are mainly caused by lacking of samples.
\end{itemize}

\subsubsection{Re-weight}
The leaf nodes are listed in TABLE \ref{tabreweight} and we ignore the nodes with no samples. Only 2 nodes are statistical significant (marked as blue) and their score difference is low. Most of the leaf nodes are effected by the limited target dataset and result in high score differences (marked as red). It's a wise choice to count more on the reasonable source model in this situation.

\begin{table*}[htbp]
  \centering
  \caption{Re-split Statistics of Intermediate Nodes}
    \begin{tabular}{ll|llll|llll|l}
    \toprule
    \textbf{node\_id} & \textbf{feat\_id} & \textbf{split\_val$_{s}$} & \textbf{split\_gain$_{s}$} & \textbf{inst\#$_{s}$} & \textbf{score$_{s}$} & \textbf{split\_val$_{t}$} & \textbf{split\_gain$_{t}$} & \textbf{inst\#$_{t}$} & \textbf{score$_{t}$} & \textbf{$\Delta$score} \\
    \midrule
    0     & 5     & 140689.5 & 542.19208 & 62400 & -0.18799 & \textcolor[rgb]{ .439,  .188,  .627}{\textbf{439902.5}} & 46.88963 & \textcolor[rgb]{ 0,  .69,  .314}{10200} & -0.18793 & \textcolor[rgb]{ 0,  .69,  .314}{6.1646E-05} \\
    1     & 43    & 11.5  & 136.56732 & 61119 & -0.19069 & 61.5  & 12.62515 & \textcolor[rgb]{ 0,  .69,  .314}{10137} & -0.18900 & \textcolor[rgb]{ 0,  .69,  .314}{0.001692} \\
    7     & 34    & 0.23066 & 46.87704 & 9607  & -0.17848 & 0.60413 & 10.84173 & \textcolor[rgb]{ 0,  .69,  .314}{3464} & -0.18281 & \textcolor[rgb]{ 0,  .69,  .314}{0.004331} \\
    15    & 41    & 6860329 & 5.63867 & 5475  & -0.19095 & 72    & 11.60792 & \textcolor[rgb]{ 0,  .69,  .314}{3445} & -0.18365 & \textcolor[rgb]{ 0,  .69,  .314}{0.007300} \\
    3     & 17    & 4318  & 136.20298 & 10933 & -0.17016 & 30515.5 & 37.96496 & \textcolor[rgb]{ 0,  .69,  .314}{3635} & -0.17824 & \textcolor[rgb]{ 0,  .69,  .314}{0.008079} \\
    24    & 40    & 0.14558 & 4.19831 & 67    & 0.10423 & 0.31587 & 4.07152 & 25    & 0.11724 & 0.013016 \\
    5     & 45    & 26.5  & 170.83625 & 650   & 0.05076 & 4.5   & 2.56277 & 40    & 0.07273 & 0.021963 \\
    8     & 34    & 0.23045 & 149.07564 & 1326  & -0.10947 & 0.50828 & 14.16933 & 171   & -0.08343 & 0.026045 \\
    2     & 43    & 0.5   & 398.10758 & 1281  & -0.05899 & 4     & 21.77838 & 63    & -0.01493 & 0.044063 \\
    17    & 42    & 834.5 & 9.89684 & 712   & -0.17151 & 1.5   & 25.89931 & 163   & -0.09701 & 0.074502 \\
    23    & 10    & 5.5   & 27.48077 & 261   & -0.09132 & 2.5   & 0.43209 & \textcolor[rgb]{ 0,  .69,  .941}{13} & 0.01176 & \textcolor[rgb]{ 0,  .69,  .941}{0.103085} \\
    16    & 5     & 17783.5 & 61.63551 & 4132  & -0.16180 & 26892.5 & 0.32298 & \textcolor[rgb]{ 0,  .69,  .941}{19} & -0.02609 & \textcolor[rgb]{ 0,  .69,  .941}{0.135712} \\
    11    & 5     & 770377.5 & 53.27773 & 328   & -0.05060 & 642766 & 2.31006 & \textcolor[rgb]{ 0,  .69,  .941}{38} & 0.08571 & \textcolor[rgb]{ 0,  .69,  .941}{0.136317} \\
    30    & 36    & -0.44286 & 5.05051 & 12    & 0.00000 & -0.46204 & 2.42667 & \textcolor[rgb]{ 0,  .69,  .941}{21} & -0.15200 & \textcolor[rgb]{ 0,  .69,  .941}{0.152000} \\
    18    & 45    & 77.5  & 93.29868 & 614   & -0.03689 & 12    & \textcolor[rgb]{ 1,  0,  0}{-0.67879} & 8     & 0.13333 & 0.170227 \\
    14    & 34    & 0.47958 & 10.55379 & 516   & -0.18538 & 0.06995 & \textcolor[rgb]{ 1,  0,  0}{-0.74462} & 22    & -0.15385 & 0.031538 \\
    6     & 43    & 23.5  & 12.37396 & 631   & -0.17165 & 25.5  & \textcolor[rgb]{ 1,  0,  0}{-0.74872} & 23    & -0.15556 & 0.016098 \\
    19    & 0     & 73687 & 5.37106 & 48096 & -0.19672 & 5148309 & \textcolor[rgb]{ 1,  0,  0}{-0.79935} & 6500  & -0.19490 & 0.001828 \\
    9     & 21    & 281.5 & 10.05906 & 48572 & -0.19640 & 29182 & \textcolor[rgb]{ 1,  0,  0}{-0.79935} & 6501  & -0.19490 & 0.001505 \\
    4     & 17    & 16899 & 57.01518 & 50186 & -0.19515 & 5454684 & \textcolor[rgb]{ 1,  0,  0}{-0.79935} & 6502  & -0.19490 & 0.000249 \\
    \bottomrule
    \end{tabular}%
  \label{tabresplit}%
\end{table*}%

\begin{table}[htbp]
  \centering
  \caption{Re-weight statistics of leaf nodes}
    \begin{tabular}{l|ll|ll|l}
    \toprule
    \textbf{node\_id} & \textbf{inst\#$_{s}$} & \textbf{score$_{s}$} & \textbf{inst\#$_{t}$} & \textbf{score$_{t}$} & \textbf{$\Delta$score} \\
    \midrule
    39    & 48067 & -0.19678 & \textcolor[rgb]{ 0,  .69,  .941}{6499} & -0.19489 & 0.00189 \\
    32    & 47    & -0.10588 & \textcolor[rgb]{ 0,  .69,  .941}{3154} & -0.18746 & 0.08158 \\
    31    & 5428  & -0.19161 & 291   & -0.14034 & 0.05127 \\
    36    & 5     & 0.06667 & 141   & -0.12828 & \textcolor[rgb]{ 1,  0,  0}{0.19494} \\
    49    & 21    & 0.02400 & 23    & 0.14074 & \textcolor[rgb]{ 1,  0,  0}{0.11674} \\
    35    & 707   & -0.17356 & 22    & 0.09231 & \textcolor[rgb]{ 1,  0,  0}{0.26587} \\
    62    & 7     & -0.09091 & 20    & -0.16667 & 0.07576 \\
    33    & 3739  & -0.16976 & 17    & -0.00952 & \textcolor[rgb]{ 1,  0,  0}{0.16023} \\
    37    & 490   & -0.07611 & 7     & 0.12727 & \textcolor[rgb]{ 1,  0,  0}{0.20339} \\
    48    & 137   & -0.15177 & 7     & -0.01818 & \textcolor[rgb]{ 1,  0,  0}{0.13359} \\
    47    & 124   & -0.02187 & 6     & 0.04  & 0.06187 \\
    50    & 46    & 0.13600 & 2     & -0.06667 & \textcolor[rgb]{ 1,  0,  0}{0.20267} \\
    34    & 393   & -0.08514 & 2     & -0.06667 & 0.01847 \\
    38    & 124   & 0.11563 & 1     & 0.04  & 0.07563 \\
    61    & 5     & 0.11111 & 1     & 0.04  & 0.07111 \\
    40    & 29    & -0.09091 & 1     & -0.04 & 0.05091 \\
    \bottomrule
    \end{tabular}%
  \label{tabreweight}%
\end{table}%

\subsubsection{Prune or Discount}
If we prune the rare branches and intermediate nodes with no gains, only 15 nodes are remained. The source patterns cannot take effect beyond doubt and the general metrics also prove this. So the discount strategy beats the prune strategy and works out better on the rare branches.

According to the above analysis, the revise strategies are determined for current scene. Both of the re-split and re-weight operations should be executed. The selection of discount value is not sensitive as long as it can differentiate the inherited source model weight and revised weight. Empirically, we set the discount value as 0.1.

\subsection{Result}
Besides the 2 baseline models, two revise mechanism mentioned in \ref{subsecmodbase} are tested, denoted as OR (OneRound) and MR (MultiRound). 

We tune the models by grid search (the tree depth in target domain is fixed as 5):
\begin{itemize}
\item tree depth in source domain (denoted as dep in the result table): 3, 4, 5
\item tree number in source domain: 2, 4, 6, 8, 10, 20, 30, 40, 60, 80
\item tree number in target domain: 40, 80, 120, 160, 200, 240
\end{itemize}

For each paramater combination, the experiment was repeated 5 times and the average value was recorded. TABLE \ref{tabresult} shows the result of the models.
\begin{table}[htbp]
  \centering
  \caption{The experiment result}
    \begin{tabular}{lllll}
    \toprule
    \textbf{Model} & \textbf{AUC} & \textbf{1\textpertenthousand\ Recall} & \textbf{AUC Lift} & \textbf{1\textpertenthousand\ Recall Lift} \\
    \midrule
    BM1   & 0.92535 & 0.16092 & -     & - \\
    \midrule
    BM2-dep3 & 0.92951 & 0.17672 & 0.449\% & 9.821\% \\
    BM2-dep4 & 0.93057 & 0.17098 & 0.563\% & 6.250\% \\
    BM2-dep5 & 0.92825 & 0.16810 & 0.313\% & 4.464\% \\
    \midrule
    OR-dep3 & 0.92886 & 0.17098 & 0.379\% & 6.250\% \\
    OR-dep4 & 0.92799 & 0.17098 & 0.284\% & 6.250\% \\
    OR-dep5 & 0.92869 & 0.17960 & 0.360\% & \textbf{11.607\%} \\
    \midrule
    MR\_dep3 & 0.93069 & 0.17241 & 0.577\% & 7.143\% \\
    MR\_dep4 & 0.92855 & 0.17672 & 0.345\% & 9.821\% \\
    MR\_dep5 & 0.92999 & 0.17098 & 0.501\% & 6.250\% \\
    \bottomrule
    \end{tabular}%
  \label{tabresult}%
\end{table}%

\begin{itemize}
\item All the transfer models promote the model performance on this dataset. Less than 10 trees are needed to reach the promotion.
\item The AUCs are high and close due to the low bad rate. OneRound mechanism offers the best 1\textpertenthousand\ recall value  and lift the BM1 by 11.607\%. The multi-round mechanism may fail to capture more source pattern information with frequent revisions.
\item The BM2 gets a better result with shallower source tree depth. In this way, it keeps the patterns to be more generic. With revise operatioins, our models make reasonable changes and remain more informative source patterns.
\end{itemize}

\subsection{Deployment}
We deploy the OneRound revision algorithm as a toolkit in PAI (Platform of Artificial Intelligence)\footnote{https://www.alibabacloud.com/press-room/alibaba-cloud-announces-machine-learning-platform-pai}, which is a machine learning platform of Alibaba Group and contains a lot of large-scale data mining algorithms.

In order to access the models in both domains, PAI offers a model pool via a specific OSS(Alibaba Cloud Object Storage Service)\footnote{https://www.alibabacloud.com/help/doc-detail/31817.htm}, which is a cost-effective, highly secure, and highly reliable cloud storage solution. 

According to the phases needed, we implement three new components: 
\begin{itemize}
\item TLXGB-SRC-TRAIN: Call the XGBoost API to train a batch of trees and push the model to the model pool (record the returned model path for the following process).
\item TLXGB-TGT-TRAIN: Fetch the source model from model pool, revise \& continue train the model and push the model to the model pool(record model path)
\item TLXGB-PREDICT: Load a model from model pool and call the XGBoost API to predict the samples
\end{itemize}

\begin{figure}[htbp]
 \centering
 \includegraphics[width=0.5\textwidth]{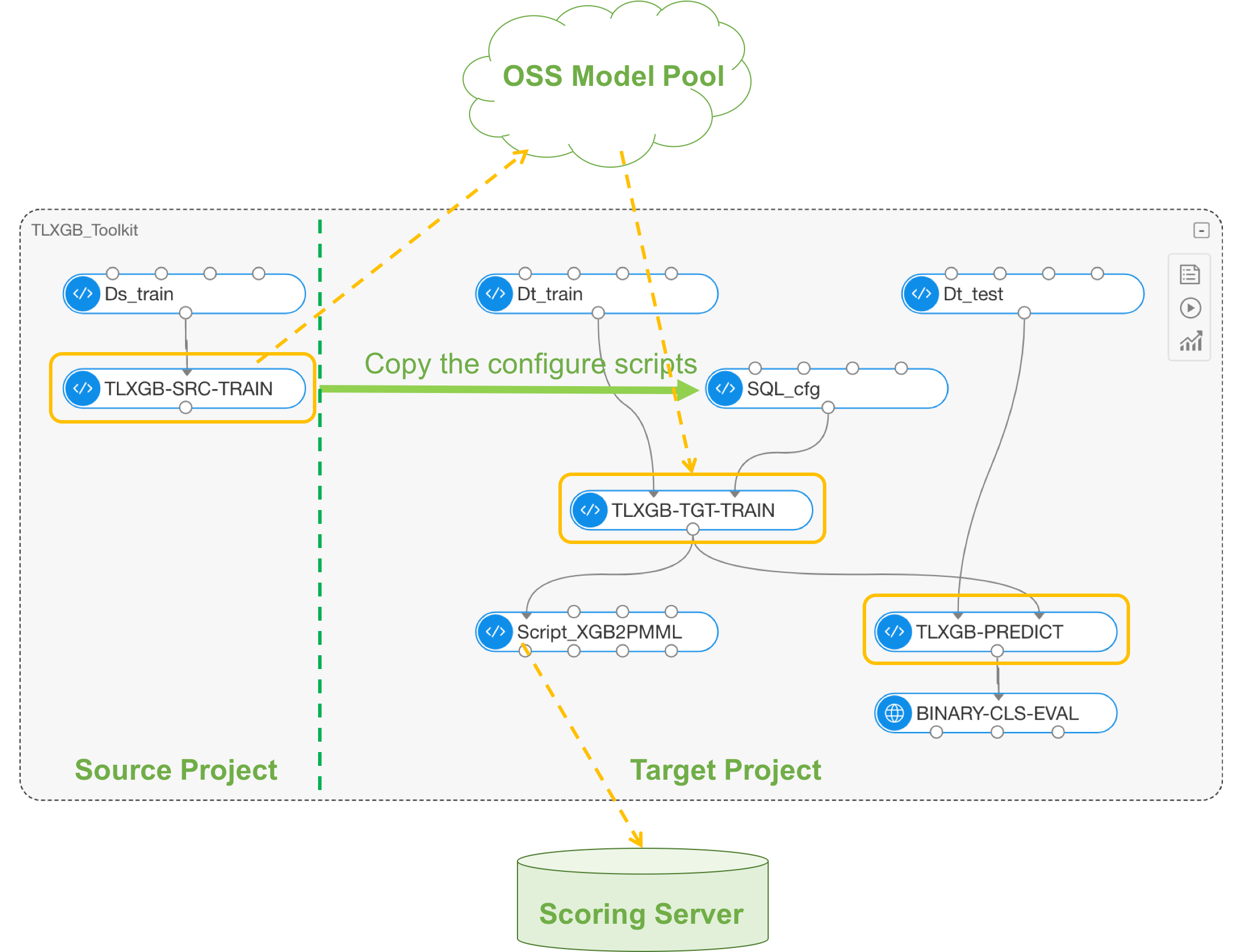}
 \caption{Workflow of the TLXGB toolkit}\label{figcomponent}
\end{figure}

Fig.\ref{figcomponent} shows the workflow of the components. Only the TLXGB-SRC-TRAIN runs on the source domain and we have to change the running project to target domain after that. Because of the data isolation, TLXGB-TGT-TRAIN cannot read the output table on source domain. We add a reconstruction SQL to the output table, copy it to a SQL component in the target project and reconstruct the configs(e.g. model path). Then continue the process with TLXGB-TGT-TRAIN and TLXGB-PREDICT. Finally, we can connect the prediction result to a evaluation component BINARY-CLS-EVAL and get the performance metrics.

The online prediction service is provided by the real-time scoring server, whose service is triggered by each payment and output a final score with the corresponding features fed. Timeliness is import for the server and the whole process is limited to 100ms.
JPMML\footnote{https://github.com/jpmml/jpmml-evaluator} is a widespread evaluator and is embeded in the server. The PMML (Predictive Model Markup Language)\footnote{http://dmg.org/pmml/v4-4/GeneralStructure.html}, an XML-based format, provides a way for analytic applications to describe and exchange predictive models produced by data mining and machine learning algorithms. 

We apply the toolkit to the card-stolen fraud detection scene. Similary to the experimental dataset, the algorithm promotes the 1\textpertenthousand\ recall impressively by 26.7\%. In order to deploy the final model in the scoring server, we wrap the JPMML-XGBoost\footnote{https://github.com/jpmml/jpmml-xgboost} and some post-process scripts to convert XGBoost models to PMML.

\section{Conclusion}\label{seccon}
The cold-starting problem under a new scene and data-sharing problem with privacy concerns can be solved by transfer learning algorithms. To deal with a real-world fraud detection modeling in Alipay, we proposed a set of revise strategies to expand the XGBoost as a transfer learning framework. Detailed analysis of the strategies are displayed. The experimental result prove it to be practical and this method promotes the model performance at the top. This model-based transfer learning framework is further deployed on PAI and work as an internal shared component. 

There are several open issues to be investigated. A practical extension is allowing the domains to maintain different feature spaces. The process for rare branches also need futher exploration and find a better way to cope with the unconfident subtrees. The revise associated parameters control the revise strategies and the parameter values rely on the data distribution. How to determine the parameter automatically is worthwhile studying. Furthermore, if the datasets are equivalent and how to conduct a multi-task learning among the domains, which is a typical scenario in private data-sharing modeling.

\bibliographystyle{IEEEtran}
\bibliography{treetl}

\begin{thebibliography}{10}
\providecommand{\url}[1]{#1}
\csname url@samestyle\endcsname
\providecommand{\newblock}{\relax}
\providecommand{\bibinfo}[2]{#2}
\providecommand{\BIBentrySTDinterwordspacing}{\spaceskip=0pt\relax}
\providecommand{\BIBentryALTinterwordstretchfactor}{4}
\providecommand{\BIBentryALTinterwordspacing}{\spaceskip=\fontdimen2\font plus
\BIBentryALTinterwordstretchfactor\fontdimen3\font minus
  \fontdimen4\font\relax}
\providecommand{\BIBforeignlanguage}[2]{{%
\expandafter\ifx\csname l@#1\endcsname\relax
\typeout{** WARNING: IEEEtran.bst: No hyphenation pattern has been}%
\typeout{** loaded for the language `#1'. Using the pattern for}%
\typeout{** the default language instead.}%
\else
\language=\csname l@#1\endcsname
\fi
#2}}
\providecommand{\BIBdecl}{\relax}
\BIBdecl

\bibitem{zhang2019distributed}
Y.-L. Zhang \emph{et~al.}, ``Distributed deep forest and its application to
  automatic detection of cash-out fraud,'' \emph{ACM Transactions on
  Intelligent Systems and Technology (TIST)}, vol.~10, no.~5, pp. 1--19, 2019.

\bibitem{pan2009survey}
S.~J. Pan and Q.~Yang, ``A survey on transfer learning,'' \emph{IEEE
  Transactions on knowledge and data engineering}, vol.~22, no.~10, pp.
  1345--1359, 2009.

\bibitem{khan2016adapting}
M.~N.~A. Khan and D.~R. Heisterkamp, ``Adapting instance weights for
  unsupervised domain adaptation using quadratic mutual information and
  subspace learning,'' in \emph{2016 23rd International Conference on Pattern
  Recognition (ICPR)}.\hskip 1em plus 0.5em minus 0.4em\relax IEEE, 2016, pp.
  1560--1565.

\bibitem{zadrozny2004learning}
B.~Zadrozny, ``Learning and evaluating classifiers under sample selection
  bias,'' in \emph{Proceedings of the twenty-first international conference on
  Machine learning}.\hskip 1em plus 0.5em minus 0.4em\relax ACM, 2004, p. 114.

\bibitem{cortes2008sample}
C.~Cortes, M.~Mohri, M.~Riley, and A.~Rostamizadeh, ``Sample selection bias
  correction theory,'' in \emph{International conference on algorithmic
  learning theory}.\hskip 1em plus 0.5em minus 0.4em\relax Springer, 2008, pp.
  38--53.

\bibitem{tan2017distant}
B.~Tan, Y.~Zhang, S.~J. Pan, and Q.~Yang, ``Distant domain transfer learning,''
  in \emph{Thirty-First AAAI Conference on Artificial Intelligence}, 2017.

\bibitem{liu2011cross}
J.~Liu, M.~Shah, B.~Kuipers, and S.~Savarese, ``Cross-view action recognition
  via view knowledge transfer,'' in \emph{CVPR 2011}.\hskip 1em plus 0.5em
  minus 0.4em\relax IEEE, 2011, pp. 3209--3216.

\bibitem{zheng2008transferring}
V.~W. Zheng, S.~J. Pan, Q.~Yang, and J.~J. Pan, ``Transferring multi-device
  localization models using latent multi-task learning.'' in \emph{AAAI},
  vol.~8, 2008, pp. 1427--1432.

\bibitem{hu2011transfer}
D.~H. Hu and Q.~Yang, ``Transfer learning for activity recognition via sensor
  mapping,'' in \emph{Twenty-second international joint conference on
  artificial intelligence}, 2011.

\bibitem{long2014transfer}
M.~Long, J.~Wang, G.~Ding, J.~Sun, and P.~S. Yu, ``Transfer joint matching for
  unsupervised domain adaptation,'' in \emph{Proceedings of the IEEE conference
  on computer vision and pattern recognition}, 2014, pp. 1410--1417.

\bibitem{duan2012domain}
L.~Duan, I.~W. Tsang, and D.~Xu, ``Domain transfer multiple kernel learning,''
  \emph{IEEE Transactions on Pattern Analysis and Machine Intelligence},
  vol.~34, no.~3, pp. 465--479, 2012.

\bibitem{blitzer2006domain}
J.~Blitzer, R.~McDonald, and F.~Pereira, ``Domain adaptation with structural
  correspondence learning,'' in \emph{Proceedings of the 2006 conference on
  empirical methods in natural language processing}.\hskip 1em plus 0.5em minus
  0.4em\relax Association for Computational Linguistics, 2006, pp. 120--128.

\bibitem{mihalkova2007mapping}
L.~Mihalkova, T.~Huynh, and R.~J. Mooney, ``Mapping and revising markov logic
  networks for transfer learning,'' in \emph{Aaai}, vol.~7, 2007, pp. 608--614.

\bibitem{mihalkova2008transfer}
L.~Mihalkova and R.~J. Mooney, ``Transfer learning by mapping with minimal
  target data,'' in \emph{Proceedings of the AAAI-08 workshop on transfer
  learning for complex tasks}, 2008.

\bibitem{davis2009deep}
J.~Davis and P.~Domingos, ``Deep transfer via second-order markov logic,'' in
  \emph{Proceedings of the 26th annual international conference on machine
  learning}.\hskip 1em plus 0.5em minus 0.4em\relax ACM, 2009, pp. 217--224.

\bibitem{ge2013oms}
L.~Ge, J.~Gao, and A.~Zhang, ``Oms-tl: a framework of online multiple source
  transfer learning,'' in \emph{Proceedings of the 22nd ACM international
  conference on Information \& Knowledge Management}.\hskip 1em plus 0.5em
  minus 0.4em\relax ACM, 2013, pp. 2423--2428.

\bibitem{zhao2011cross}
Z.~Zhao, Y.~Chen, J.~Liu, Z.~Shen, and M.~Liu, ``Cross-people mobile-phone
  based activity recognition,'' in \emph{Twenty-second international joint
  conference on artificial intelligence}, 2011.

\bibitem{pan2008transferring}
S.~J. Pan, D.~Shen, Q.~Yang, and J.~T. Kwok, ``Transferring localization models
  across space.'' in \emph{AAAI}, 2008, pp. 1383--1388.

\bibitem{Kienzle2006Personalized}
W.~Kienzle and K.~Chellapilla, ``Personalized handwriting recognition via
  biased regularization,'' in \emph{International Conference on Machine
  Learning}, 2006.

\bibitem{rodner2008learning}
E.~Rodner and J.~Denzler, ``Learning with few examples using a constrained
  gaussian prior on randomized trees.'' in \emph{VMV}, 2008.

\bibitem{tommasi2010safety}
T.~Tommasi, F.~Orabona, and B.~Caputo, ``Safety in numbers: Learning categories
  from few examples with multi model knowledge transfer,'' in \emph{2010 IEEE
  Computer Society Conference on Computer Vision and Pattern
  Recognition}.\hskip 1em plus 0.5em minus 0.4em\relax IEEE, 2010, pp.
  3081--3088.

\bibitem{rodner2011learning}
E.~Rodner and J.~Denzler, ``Learning with few examples for binary and
  multiclass classification using regularization of randomized trees,''
  \emph{Pattern Recognition Letters}, vol.~32, no.~2, pp. 244--251, 2011.

\bibitem{rettinger2006boosting}
A.~Rettinger, M.~Zinkevich, and M.~Bowling, ``Boosting expert ensembles for
  rapid concept recall,'' in \emph{Proceedings of the National Conference on
  Artificial Intelligence}, vol.~21, no.~1.\hskip 1em plus 0.5em minus
  0.4em\relax Menlo Park, CA; Cambridge, MA; London; AAAI Press; MIT Press;
  1999, 2006, p. 464.

\bibitem{luo2008transfer}
P.~Luo, F.~Zhuang, H.~Xiong, Y.~Xiong, and Q.~He, ``Transfer learning from
  multiple source domains via consensus regularization,'' in \emph{Proceedings
  of the 17th ACM conference on Information and knowledge management}.\hskip
  1em plus 0.5em minus 0.4em\relax ACM, 2008, pp. 103--112.

\bibitem{ruckert2008kernel}
U.~R{\"u}ckert and S.~Kramer, ``Kernel-based inductive transfer,'' in
  \emph{Joint European Conference on Machine Learning and Knowledge Discovery
  in Databases}.\hskip 1em plus 0.5em minus 0.4em\relax Springer, 2008, pp.
  220--233.

\bibitem{baxter2000model}
J.~Baxter, ``A model of inductive bias learning,'' \emph{Journal of artificial
  intelligence research}, vol.~12, pp. 149--198, 2000.

\bibitem{pratt1991direct}
L.~Y. Pratt, J.~Mostow, C.~A. Kamm, and A.~A. Kamm, ``Direct transfer of
  learned information among neural networks.'' in \emph{AAAI}, vol.~91, 1991,
  pp. 584--589.

\bibitem{thrun1994learning}
S.~Thrun and T.~M. Mitchell, ``Learning one more thing,'' CARNEGIE-MELLON UNIV
  PITTSBURGH PA DEPT OF COMPUTER SCIENCE, Tech. Rep., 1994.

\bibitem{eaton2008modeling}
E.~Eaton, T.~Lane \emph{et~al.}, ``Modeling transfer relationships between
  learning tasks for improved inductive transfer,'' in \emph{Joint European
  Conference on Machine Learning and Knowledge Discovery in Databases}.\hskip
  1em plus 0.5em minus 0.4em\relax Springer, 2008, pp. 317--332.

\bibitem{Deng2014Cross}
W.~Y. Deng, Q.~H. Zheng, and Z.~M. Wang, ``Cross-person activity recognition
  using reduced kernel extreme learning machine,'' \emph{Neural Networks},
  vol.~53, no.~5, pp. 1--7, 2014.

\bibitem{Yang2007Cross}
J.~Yang, Y.~Rong, and A.~G. Hauptmann, ``Cross-domain video concept detection
  using adaptive svms,'' 2007.

\bibitem{Duan2009Domain}
L.~Duan, I.~W. Tsang, D.~Xu, and T.~S. Chua, ``Domain adaptation from multiple
  sources via auxiliary classifiers,'' in \emph{International Conference on
  Machine Learning}, 2009.

\bibitem{long2015learning}
M.~Long, Y.~Cao, J.~Wang, and M.~I. Jordan, ``Learning transferable features
  with deep adaptation networks,'' \emph{arXiv preprint arXiv:1502.02791},
  2015.

\bibitem{long2016deep}
M.~Long, J.~Wang, Y.~Cao, J.~Sun, and S.~Y. Philip, ``Deep learning of
  transferable representation for scalable domain adaptation,'' \emph{IEEE
  Transactions on Knowledge and Data Engineering}, vol.~28, no.~8, pp.
  2027--2040, 2016.

\bibitem{long2017deep}
M.~Long, H.~Zhu, J.~Wang, and M.~I. Jordan, ``Deep transfer learning with joint
  adaptation networks,'' in \emph{Proceedings of the 34th International
  Conference on Machine Learning-Volume 70}.\hskip 1em plus 0.5em minus
  0.4em\relax JMLR. org, 2017, pp. 2208--2217.

\bibitem{Tzeng2015Simultaneous}
E.~Tzeng, J.~Hoffman, T.~Darrell, and K.~Saenko, ``Simultaneous deep transfer
  across domains and tasks,'' 2015.

\bibitem{domingos2000mining}
P.~Domingos and G.~Hulten, ``Mining high-speed data streams,'' in \emph{Kdd},
  vol.~2, 2000, p.~4.

\bibitem{hulten2001mining}
G.~Hulten, L.~Spencer, and P.~Domingos, ``Mining time-changing data streams,''
  in \emph{Proceedings of the seventh ACM SIGKDD}.\hskip 1em plus 0.5em minus
  0.4em\relax ACM, 2001, pp. 97--106.

\bibitem{jin2003efficient}
R.~Jin and G.~Agrawal, ``Efficient decision tree construction on streaming
  data,'' in \emph{Proceedings of the ninth ACM SIGKDD}.\hskip 1em plus 0.5em
  minus 0.4em\relax ACM, 2003, pp. 571--576.

\bibitem{nunez2007learning}
M.~N{\'u}{\~n}ez, R.~Fidalgo, and R.~Morales, ``Learning in environments with
  unknown dynamics: Towards more robust concept learners,'' \emph{Journal of
  Machine Learning Research}, vol.~8, no. Nov, pp. 2595--2628, 2007.

\bibitem{segev2016learn}
Segev \emph{et~al.}, ``Learn on source, refine on target: a model transfer
  learning framework with random forests,'' \emph{IEEE transactions on pattern
  analysis and machine intelligence}, vol.~39, no.~9, pp. 1811--1824, 2016.

\bibitem{dai2007boosting}
W.~Dai, Q.~Yang, G.-R. Xue, and Y.~Yu, ``Boosting for transfer learning,'' in
  \emph{Proceedings of the 24th international conference on Machine
  learning}.\hskip 1em plus 0.5em minus 0.4em\relax ACM, 2007, pp. 193--200.

\bibitem{freund1997decision}
Y.~Freund and R.~E. Schapire, ``A decision-theoretic generalization of on-line
  learning and an application to boosting,'' \emph{Journal of computer and
  system sciences}, vol.~55, no.~1, pp. 119--139, 1997.

\bibitem{friedman2001greedy}
J.~H. Friedman, ``Greedy function approximation: a gradient boosting machine,''
  \emph{Annals of statistics}, pp. 1189--1232, 2001.

\bibitem{chen2016xgboost}
T.~Chen and C.~Guestrin, ``Xgboost: A scalable tree boosting system,'' in
  \emph{Proceedings of the 22nd acm sigkdd international conference on
  knowledge discovery and data mining}.\hskip 1em plus 0.5em minus 0.4em\relax
  ACM, 2016, pp. 785--794.

\bibitem{zhou2017psmart}
J.~Zhou \emph{et~al.}, ``Psmart: Parameter server based multiple additive
  regression trees system,'' in \emph{Proceedings of the 26th International
  Conference on World Wide Web Companion}, 2017, pp. 879--880.

\bibitem{fang2018unpack}
W.~Fang, J.~Zhou, X.~Li, and K.~Q. Zhu, ``Unpack local model interpretation for
  gbdt,'' in \emph{International Conference on Database Systems for Advanced
  Applications}.\hskip 1em plus 0.5em minus 0.4em\relax Springer, 2018, pp.
  764--775.

\end{thebibliography}

\end{document}